Limited Liability Partnership

"Astana IT University"

**STUDENT SCIENTIFIC-RESEARCH WORK**

**Topic: Developing a Dataset-Adaptive, Normalized Metric for Machine Learning Model Assessment: Integrating Size, Complexity, and Class Imbalance**

Completed by:

Ossenov Serzhan,

Big Data Analysis, BDA-2207

Supervisor:

Aidana Zhalgas,

MSc, Associate professor

Astana, 2024

Contents

**Introduction**

There exists a great variety of metrics to track performance of the machine learning algorithms, including those designed for classification, clustering and regression tasks. Some examples of the most used of them are accuracy, precision and F1-score. These metrics are more prevalent for evaluation of classical machine learning models such as support vector machine, logistic regression, decision tree and others, because they are easily interpretable and directly address specific task's need. They can also be exploited for deep learning models, but other metrics are preferable for neural networks, because of more complex scenarios, where models encounter a large number of classes and extremely high features dimensionality. According to Brenton Adey [1], accuracy almost always tends to grow with the increase of dataset's size. With growth of available data, machine learning models are able to find more patterns and generalize better on the content of dataset. Feeding a model with small dataset usually results in low accuracy, meanwhile high accuracy arises from feeding the same model with large dataset. However, there is limited literature discussing normalized metrics designed to assess a model's potential based on its performance based on several dataset properties such as size, features dimensionality and class imbalances. By incorporating these properties into single metric, the potential upper limits of the model's performance could be revealed. This metric would provide a relative evaluation that highlights how well the model would perform if the dataset were larger or had various complexities. This approach contrasts drastically from established metrics that rely solely on raw accuracy, as it aims to provide more scalable and flexible evaluation of model potential under conditions of data scarcity and poor quality. Introduction of such normalized metric would save up computational resources during the construction of model itself. Research performed by Simona Maggio [2] states that hyperparameter search becomes tedious with large datasets. It involves training hundreds or thousands of individual models, each having different parameters, on the training dataset. By that, this approach finds the best set of features and their values based on accuracy or other metrics obtained after each training process. However, it uses too much computational resources. Normalized metric would save resources, as hyperparameter search would be executed on a small proportion of the original dataset. Minimum number of samples are enough for this metric to evaluate the model's potential effectiveness considering all of dataset's unfavorable conditions if they were present. Consequently, this metric would quickly indicate whether a model is well-constructed without the need to exhaustively train and compare thousands of models on the full dataset. Secondly, such adjusted measures would help guard against a false sense of optimality by providing a more realistic assessment of model performance. According to Michael Lones [3], researchers are prone to believe their model has achieved optimal performance when it actually has not. A normalized metric would boost the accuracy of models trained under challenging conditions but penalize those that perform only moderately on large datasets. For instance, practitioners may consider 85% accuracy acceptable; however, if it was achieved under favorable conditions - large dataset size, low feature dimensionality, minimal class imbalance—this metric would indicate potential for improvement. This approach opens up prospects for further improvement by eliminating the false sense of optimality, therefore researchers seeking optimal results will not stop at such a mediocre result. In addition, the metric described would facilitate model evaluation in the context of limited data in the domain. As stated in research performed by Georgios Douzas [4], in the early phases of product development, manufacturing companies typically deal with a small number of samples, whereas healthcare organizations must deal with many rare disorders for which there are few records available. The popular UCI machine learning repository contains some datasets from the healthcare domain. To be precise, the Breast Cancer dataset has only 699 samples, whereas only 583 samples are available in the Indian Liver Patient dataset. For this reason, introducing a normalized metric that assesses a model's potential with a small dataset allows practitioners to develop effective predictive models during the early stages. This prepares the model to scale and adapt seamlessly when larger datasets become available, ensuring it performs well as more data is acquired. The metric should be developed in a way that it depends only on properties that are straightforward to obtain, such as dataset size, feature dimensionality, and class imbalance ensuring it remains practical and widely applicable across a range of model types. Advanced properties, which would involve advanced preprocessing or complex analysis, are not pursued to keep the metric user-friendly and feasible for real-world applications.

Survey

A survey was conducted among students from different universities in Kazakhstan and the USA to assess the applicability and possible demand for these new prospects. The purpose of this survey is to learn more about typical problems that arise during model evaluation and programming. It will also reveal whether or not students rely on standard metrics without question and to what extent they trust them. A comprehensive list of questions about this topic, including 8 closed and 2 open-ended questions, was created to explore and reveal insights.

List of questions:

1. How often do you find traditional metrics (like accuracy, F1-score) insufficient for evaluating models on small or imbalanced datasets?

2. What dataset challenges have you encountered most frequently when training machine learning models? (Select all that apply)

3. Do you believe that model performance on small datasets could provide a reliable estimate of the model's potential on larger datasets?

4. How useful would a metric that adjusts for dataset properties (e.g., size, imbalance, dimensionality) be in your experience with machine learning projects?

5. Have you ever felt that an accuracy score gave a false sense of optimality for a model trained on favorable conditions (e.g., large dataset, balanced classes)?

6. When conducting hyperparameter tuning on large datasets, how challenging do you find it to balance computational cost with model accuracy?

7. How valuable would you find a metric that estimates a model's potential using only a small sample of the dataset, potentially saving time and computational resources?

8. Would you be likely to use a metric that allows for a "scaled" evaluation of a model, considering data scarcity and quality, to estimate its potential performance on larger datasets?

9. If you have worked on training machine learning models with challenging datasets, please describe the main limitations you encountered with traditional metrics like accuracy or F1-score when evaluating model performance on these datasets.

10. In your opinion, would a metric that adjusts for dataset properties (like size, imbalance, and dimensionality) and allows performance estimation on a smaller dataset be beneficial?

If so, explain how this could impact your approach to model evaluation and optimization, especially in terms of time and resource savings.

A total of 301 participants from the USA and Kazakhstan, who are either studying computer science or have experience with machine learning, took part in the survey.

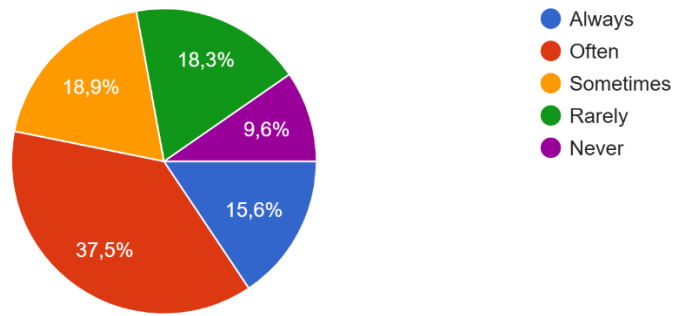

Figure 1 - Viewpoints on the sufficiency of traditional metrics for evaluating models on small or imbalanced datasets

According to the chart, 53.1% of respondents believe that traditional measures are never or rarely useful for assessing a model's performance when constructed on a poor dataset. In contrast, 27.9% of respondents stated they never or seldom doubt metrics like accuracy and F1-score as they assume metrics remain useful even when the dataset conditions are unfavorable. 18.9% of participants said that they occasionally feel these measurements lacking, indicating a balanced perspective in which traditional metrics may be helpful but may not always provide a thorough assessment.

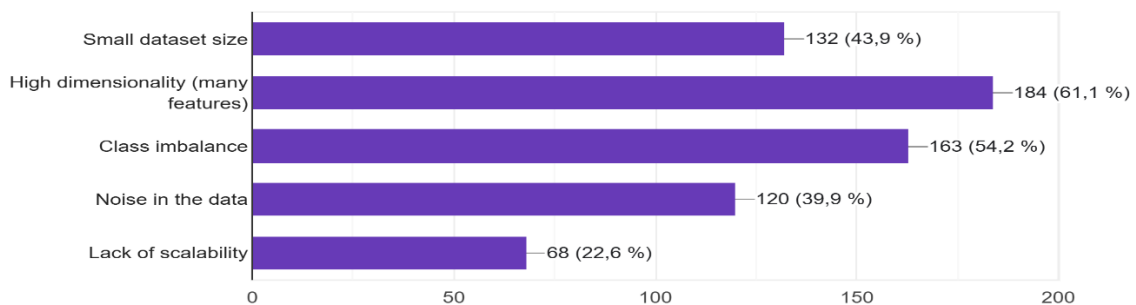

Figure 2 - The most frequent dataset issues faced during machine learning model training

Dealing with high-dimensional datasets was mentioned by 61.1% of participants as the most frequent issue, as the bar chart illustrates. 54.2% of respondents cited class imbalance as the next most common problem while training machine learning models. Another major issue that 43.9% of participants faced was a lack of data. However, just 39.9% and 22.6% of respondents, respectively, claimed that they were affected by data noise and a lack of scalability.

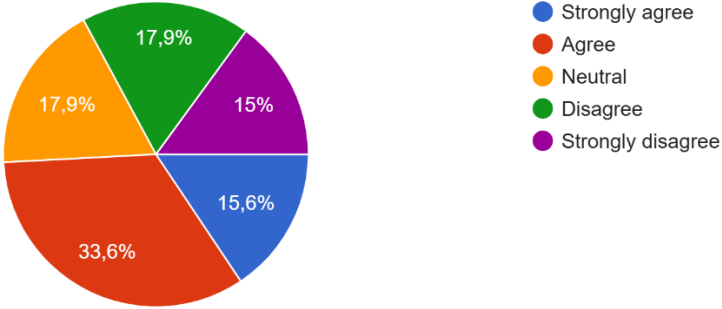

Figure 3 - Perceptions on the reliability of small dataset performance as a possible indicator for larger datasets

The pie chart illustrates that 33.6% of respondents believe model performance on small datasets can provide an early indicator of how the model would perform on larger datasets. Opinions on the other options, however, are divided almost evenly. 17.9% of respondents, for instance, disagree with the statement, and another 17.9% are neutral. Furthermore, 15% strongly disagree that the performance of small datasets alone may be used to evaluate a model's potential. On the other hand, 15.6% of respondents strongly agree with this claim.

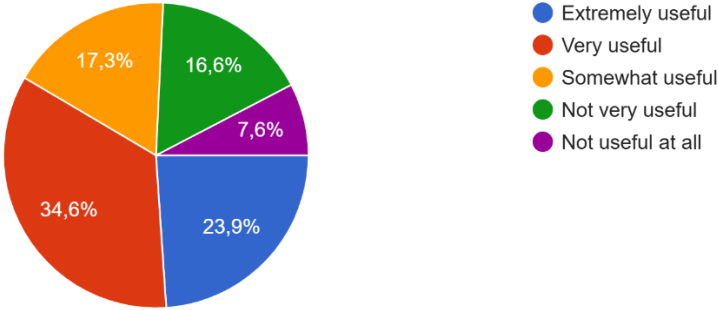

Figure 4 - Perception of dataset-adjusted metrics' usefulness in machine learning projects

According to the chart, 34.6% of respondents agree that machine learning projects would benefit greatly from an estimate that accounts for dataset characteristics like size, imbalance, and dimensionality. Furthermore, 23.9% of respondents think it is very useful, which shows that measures that take dataset characteristics into account are strongly supported. A moderate level of interest is indicated by 17.3% of participants who believe that such measurements would be somewhat helpful. Conversely, 16.6% of respondents say these metrics would be of little or no utility, while a smaller percentage of 7.6% think they would be not useful at all.

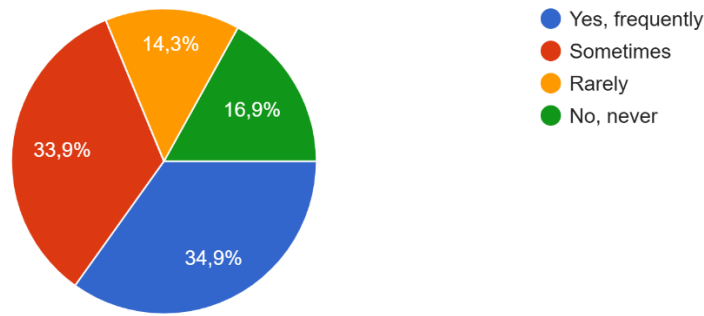

Figure 5 - Perspectives on the reliability of accuracy scores under favorable model conditions

The chart shows that 34.9% of participants commonly believed that when models were trained in favorable conditions, as with balanced classes and huge datasets, accuracy scores gave a false sense of optimality. According to 33.9% of respondents, they sometimes encountered this problem, indicating that many participants are aware of the possible drawbacks of using accuracy as the only performance metric. Less frequently, 14.3% of participants said they only occasionally encountered this issue, and 16.9% said they hadn't considered an accuracy score to be an inaccurate representation of a model's performance.

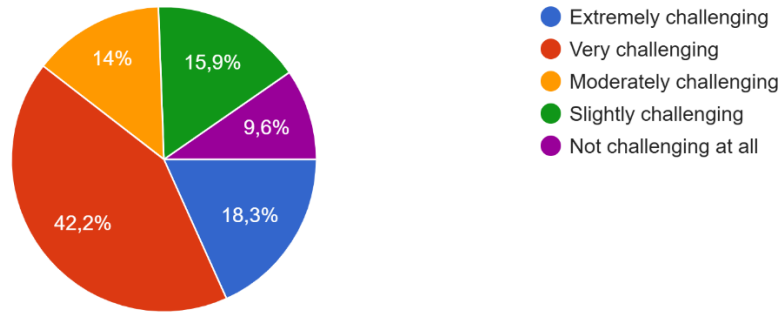

Figure 6 - Challenges in hyperparameter tuning on large datasets

According to the chart, 42.2% of respondents stated it was very difficult to balance model accuracy and computing cost while hyperparameter tuning on big datasets. This indicates that a sizable percentage of participants find this part of model training problematic. Furthermore, 18.3% of respondents reported that they find this procedure to be extremely challenging, highlighting the significant challenge that many people encounter when improving models. 15.9% of participants claimed it was slightly challenging and 14% noted it was moderately difficult. 9.6% of respondents said it is not difficult at all to achieve a balance between accuracy and computational expense.

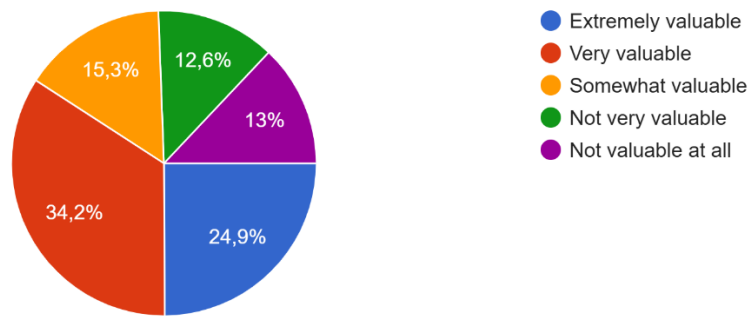

Figure 7 - Value of small-sample metrics for model potential estimation

The chart illustrates that 34.2% of respondents believe it is very valuable when a metric can predict a model's potential using a small sample of the dataset, potentially saving time and computational resources. A significant appreciation for methods that can offer early insights while preserving resources is demonstrated by 24.9% of participants who considered such a metric to be extremely valuable. Of those surveyed, 12.6% believe this kind of metric is not very valuable, while 15.3% think it is somewhat valuable. Only 13% of respondents think that this kind of measure is worthless. The perceived value is especially highlighted in the context of conserving time and computational resources, as opposed to one of the earlier charts where respondents emphasized the general utility of metrics that adapt for dataset attributes (e.g., size, imbalance, dimensionality). It shows that while efficiency-driven metrics have a lot of support, a sizable percentage of respondents are still skeptical about their applicability.

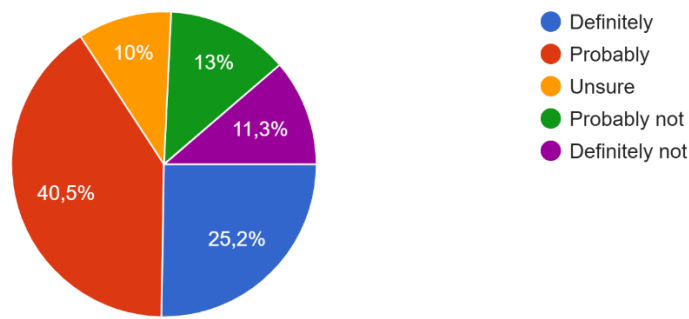

Figure 8 - Likelihood of using a scaled metric for evaluating model performance considering data scarcity and quality

According to the figure, more than one-third of respondents would probably estimate a model's performance on larger datasets using a metric that enables a "scaled" evaluation of a model, accounting for data quality and availability. About one-quarter of participants indicated that they would definitely utilize such a metric, indicating a significant willingness. 10% of participants are still uncertain about its value. A lower percentage of respondents said they would certainly not consider using such a metric, while 13% said they would probably not use it.

The last two open-ended questions are intended to gather information about the real-world difficulties and constraints people encounter when applying traditional evaluation metrics, such as accuracy and F1-score, to complex datasets. The purpose of the first question is to identify particular problems that arise when dealing with difficult datasets, such as small, imbalanced, or noisy data. It looks for thorough answers about the drawbacks of standard measures in such instances. The perceived value of a new metric that accounts for dataset characteristics and enables performance estimation on smaller samples is examined in the second question. It seeks to comprehend the potential effects of such a metric on model optimization and evaluation, particularly concerning time and resource efficiency. The following table includes some examples of participants' responses:

| Example | Focus area | Response |
| --- | --- | --- |
| Example 1 | Imbalanced Datasets | Traditional metrics like accuracy have limitations when working with imbalanced datasets. On such datasets, accuracy may appear high because the model correctly classifies the majority class while ignoring the minority. This gives a false sense of model performance, as it fails to capture the ability to recognize rare but important classes. |
| Example 2 | Complex Data Patterns | When working with challenging datasets, traditional metrics like accuracy or F1-score can be misleading. For imbalanced classes, accuracy may overestimate performance, while F1-score struggles to capture nuanced errors in complex data patterns. These metrics also lack insights into model calibration, robustness, and real-world generalizability, which are critical when datasets have noise or subtle biases. |
| Example 3 | High Dimensionality | Traditional metrics like accuracy or F1-score can provide a skewed evaluation of imbalanced or high-dimensional data. For instance, on imbalanced datasets, accuracy may be high even if the model ignores rare classes, and high dimensionality can lead to overfitting. |

Table 1 - Selected responses on limitations of traditional metrics

The next question aims to investigate participants' perceptions of the possible advantages, especially in terms of time and resource savings, of a metric that accounts for dataset characteristics and allows performance assessment on smaller datasets. On this topic, opinions differed. Positive responses highlighted possible benefits for enhancing model evaluation and optimization efficiency, while other respondents expressed doubts regarding the usefulness of such a metric. The next table contains some of the responses:

| Example | Focus area | Response |
|---|---|---|
| Example 1 | Efficiency and Resource Savings | A metric that considers dataset characteristics like size and imbalance would greatly enhance model evaluation. It would allow us to estimate performance more reliably on limited data, which is crucial in cases where data is costly or hard to obtain. This would streamline the model development process, allowing for quicker decision-making and more efficient resource usage. |
| Example 2 | Real-World Applicability and Class Imbalance | I think so. This metric would make it possible to more accurately assess the performance of models in real-world conditions, especially in cases with unbalanced classes, providing a balanced view of the model's ability to handle different classes. |
| Example 3 | Skepticism About Predictive Accuracy on Large Datasets | A metric derived from a short dataset, in my opinion, is unable to predict a model's performance on larger datasets. Although the concept seems practical, it ignores the complexity and unpredictability of actual data. A tiny sample size could produce misleading results and create a mistaken sense of confidence. |

Table 2 - Summary of responses on the usefulness of dataset-adjusted metrics

The survey gave important information on the requirements and difficulties experienced by machine learning professionals while handling complicated datasets. Traditional metrics like accuracy and F1-score were frequently cited by most respondents as having problems, particularly when working with high-dimensional or imbalanced data, data scarcity, and other difficult dataset features. Because traditional measures are unable to represent the complex performance needs of models trained under these conditions, these constraints frequently result in evaluations that are deceptive. The majority of participants were optimistic about the potential benefits of an innovative metric that accounts for the size, imbalance, and dimensionality of datasets. Many believe that even for smaller data samples, such a metric could yield more accurate performance predictions, which would be very advantageous in terms of saving time and resources. This would facilitate faster decision-making and more effective model development processes, especially in situations where data collection is costly, time-consuming, or challenging. Some respondents, however, expressed uncertainty about the feasibility and accuracy of utilizing small amounts of data to forecast model performance on larger datasets. They underlined that a tiny fraction may not adequately represent the complexity of real-world data, which could result in overconfidence or inaccurate findings regarding model capability. In conclusion, the survey shows a high level of interest in creating a dataset-aware, adaptive metric that can overcome the drawbacks of traditional evaluation metrics. More efficient model evaluation and optimization may be facilitated by such a tool, particularly for those dealing with difficult or limited data. However, any new metric would have to show reliability and consistency across different dataset kinds and settings in order to be widely utilized. The results and insights from the survey will be considered and applied in the development and testing of the normalized metric.

Summary

To sum up everything that has been stated above, this research paper aims to develop a metric which can balance practicality and effectiveness and will be used to estimate model performance in situations of limited data. It is designed to rely solely on easily obtainable properties such as dataset size, feature dimensionality, class imbalance, and others, avoiding advanced characteristics that demand complex preprocessing for the metric to remain accessible and applicable across a variety of model types. This metric additionally supports realistic evaluation standards by including a penalization feature for models that perform poorly even after training on huge datasets. By addressing the research question, "What impact does a normalized, property-based metric have on evaluating model performance under varying dataset constraints, and how effectively does it predict scalability as data availability increases?", this method opens prospects for scalable model optimization, preparing models to adapt and perform well as larger datasets become available.

**Literature Review**

Class imbalance significantly affects the reliability of accuracy as a traditional metric in machine learning because often one class significantly outweighs others. This means that in many scenarios, a model that always predicts the majority class will have a high accuracy without learning any meaningful patterns. This especially becomes a problem when the minority classes are of primary interest [5]. For example, the fraud detection problem in which only 1% of all transactions are fraudulent. The model will always predict "not fraud" with 99% accuracy, but it will fail to identify any actual fraudulent cases, hence it is practically useless. Another limitation of accuracy and other traditional metrics is that these metrics cannot take into consideration misclassification costs. It does so because all errors have equal weights. In real applications such as medical the cost of a false negative instance, classifying a sick patient as healthy can be much worse than a false positive [6]. Accuracy does not take these different misclassification costs into account; thus, it is an unsatisfactory metric for any scenario in which errors have different consequences. Various works have focused on the impact of dataset characteristics on machine learning model performance. Most of these works have targeted the influence of meta-level features, dataset size, and feature selection. Research published in Nature demonstrated how different meta-level and statistical features of tabular datasets influence various machine learning algorithms in performance [7]. These included dataset size, number of attributes, and the ratio between positive and negative class instances as some of the meta-level features analyzed, while statistical features included mean, standard deviation, skewness, and kurtosis. These features had a great effect on the performances of models, as evidenced by the results. For example, datasets with high-skewed distributions needed specialized treatments or transformations for effective learning to take place. Another recent study investigated the sensitivity of various classifiers due to dataset size, conducted by the researchers at MIT [8]. Their results showed that most of the classifiers degrade in performance as the size of the training set decreases. However, not all models were equally sensitive to dataset size: while Decision Trees were highly sensitive, Random Forest and Neural Networks showed moderate sensitivity. More importantly, in cases of smaller datasets, a well-representative sample was more important than the type of chosen classification model; it means that, for cases with some constraints, data quality was more important than model selection. Two notable works address the challenge of limited data in machine learning by showing different ways of improving model performance. The work "Training Deep Learning Models with Small Datasets" by Miguel Romero [9] covers one-cycle training, discriminative learning rates with progressive freezing, and parameter adaptation after transfer learning. These methods are of particular use in domains like medical imaging, in which large datasets are often beyond reach. The authors then demonstrate how these strategies enhance the accuracy and robustness of the models even when working with small data. In a similar way, "Making the Most of Small Software Engineering Datasets with Modern Machine Learning" by Julian Aron Prenner and Romain Robbes studies the adaptation of pre-trained Transformer models to smaller software engineering datasets [10]. This work benchmarks active learning, data augmentation, and intermediate-task fine-tuning, providing practical recommendations on how to maximize performance in resource-limited settings. However, there is still a shortage of research that comprehensively takes into account all dataset characteristics, including class imbalance, feature dimensionality, and noise, and develops a formula for a normalized metric that incorporates these factors, even though these studies offer insightful information about how to improve machine learning models under limited data. The gap emphasizes the necessity of more research to create universal measures that can handle a wider variety of data challenges.

**Methods**

Dataset size

Dataset size is a paramount component, because the more data available, the more likely machine learning model learns some patterns in this data. However, there is no universal threshold that differentiates small datasets from medium and large ones. Each situation is unique and context dependent. There are a variety of machine learning algorithms and all of them require different amounts of data to be fed with to perform well. As an example, complex deep learning algorithms need more samples to generalize well in comparison with classic models, like SVM, Logistic regression and K-means [11]. There are some heuristically obtained boundaries to distinguish between different-sized models, however all of them lack theoretical arguments. It can be explained by the Sotires paradox. Calling a dataset with one sample "small" may be objectively understandable, then adding one more sample certainly does not make it "large" Adding a third sample does not make it "large." Following this logic, at each step adding one more sample, even up to 10,000, does not mark the boundary beyond which the data set becomes "large". On the other hand, if a dataset of 10,000 samples is considered "large", taking one sample away cannot suddenly render it "small." Likewise, taking another sample away still doesn't grant it a different status. By this kind of argument, taking samples away one at a time until only a single sample remains in the dataset, based on each of these steps individually, at no stage would it be labeled "small". This demonstrates the paradox from the other end: incremental losses don't provide a clear boundary for when "large" becomes "small" [12]. Moreover, the principle of diminishing returns is another argument against incorporating dataset's size as completely separate factor in the formula of normalized metric. After a certain point, increasing the dataset size makes progressively smaller improvements in model performance. Initially, adding more data to a small dataset results in impressive efficiency gains, but at some point, machine learning models already learnt valuable patterns and expanding data size even further results in almost unnoticeable increases in accuracy. For example, it's possible that a model that has been trained with 10,000 samples currently produces results that are near the peak and that an additional 10,000 would just slightly improve the results. Large datasets may be over-punished for indicating such a significant performance improvement only because of their size, if dataset size is directly factored into the formula. A shift of focus on data quality and features would be better representative of the capability of the model rather than just its size. To provide an example, research carried out by Rhitabrat Pokharel [13] aimed to perform sentiment analysis on YouTube comments by using five different models.

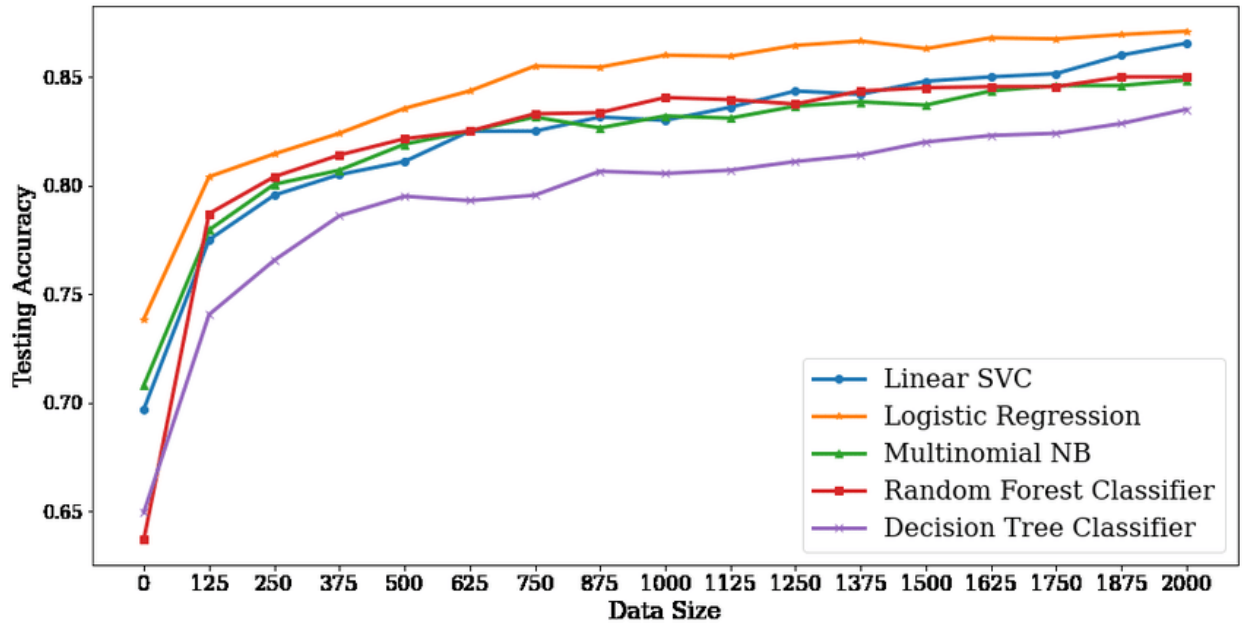

Figure 9 - Accuracy of Machine Learning Models with Increasing Data Size

Figure 9 shows that accuracy across all models rapidly rises from 0 to 125 data samples. Then, the rate of growth decreases a little from 125 to 500 data entities. Finally, there is almost no increase in performance after 500 data samples, with accuracy ranging from 0.8 to 0.85. Moreover, because each machine learning model uses a different mathematical approach to find the pattern, they all have different data requirements. Because of their lesser complexity, which results in the estimate of fewer parameters, basic models like logistic regression and linear regression would have a limited dataset, especially when the relationships are linear. Data is divided into branches via tree-based models, such as random forests and decision trees. However, in order to obtain an accurate depiction of branches, the tree may require more samples when dealing with high-dimensional data. Hyperplanes are used to divide classes into SVMs, and in order to prevent overfitting, big datasets in a high-dimensional space are required. To conclude, the metric will not include dataset size as a separate factor; instead, it will focus on important dataset characteristics that reflect its quality, with dataset size indirectly influencing the metric.

Feature Dimensionality

Nowadays, data expands in unstructured and high-dimensional ways, resulting in datasets having many characteristics that make it challenging for machine learning models to learn from. With the increase in the number of features in the dataset, it becomes harder to recognize patterns between various combinations of these features, therefore models require more data to increase the possibility of capturing the essence of tendencies in the dataset. However, large size data is limited in some domains such as bioinformatics, gene expression and data mining. As a result, it becomes challenging to develop a predictive algorithm due to high dimension and low-sample-size datasets prevalence [14]. There are many techniques to reduce dimensionality, such as PCA, t-SNE, LDA and others. However, even after applying them, there still exists a probability that many features remain, because some methods, like PCA retain the most useful characteristics. In such cases, normalized metric needs to take dataset's features dimensionality into account. There is not a fixed number that distinctly separates high-dimensional from low-dimensional datasets. Instead, dimensionality is relative and depends on the nature of data itself [15]. Nevertheless, some justified ratio is still needed to formulate a relationship between dataset's size and number of features. Research by V. Vapnik [16], exploited statistical learning theory to obtain approximate permissible ratio between number of samples in a dataset and its dimensions. According to its results, the allowable ratio is 20, meaning that to minimize overfitting, at least twenty samples should be available for each feature. The main term would be derived from this ratio - $\frac{d}{0.05N}$. As some classic metric, such as accuracy will be multiplied by feature-dimensionality related factor, it must handle all possible scenarios summarized in the table below.

|  | Small Dataset | Large Dataset |
| --- | --- | --- |
| Few features | Stable performance, reduced overfitting; dimensionality is manageable for limited data. | Optimal performance: low dimensionality combined with sample data leads to generalizability. |
| Many features | Poor performance due to overfitting risks; insufficient data to support high dimensionality. | Good performance as sufficient data supports complex feature interactions. |

Table 3 - Summary of Model Performance Across Different Dataset Sizes and Feature Dimensionalities

The main term defines the threshold between small and large datasets and features. In optimal scenario, when number of samples is exactly twenty times greater than number of features, it equals 1. For example, d = 10, n = 200, $\frac{d}{0.05N}$ = 1, therefore multiplying accuracy by that term in such case does not penalize or awards model's performance. This factor increases the accuracy when the number of features exceeds a defined limit, reflecting that the model has greater potential to perform better with the increasing size of the dataset. It is expected that the model generalizes better with more data and a consistent number of features. Thus, this metric hence measures the potential improvement of the model in case of future expansion of data. However, the result can vary significantly. To start with, expression $\frac{d}{0.05*N}$ can be rewritten as $20 * \frac{d}{N}$. This form scales the ratio $\frac{d}{N}$ by a factor of 20. By examining the behavior of $\frac{d}{N}$ under different conditions, it is possible to determine the possible the range of initial term $\frac{d}{0.05*N}$. Firstly, when $N$ grows faster than $d$, approaching $+\infty$ or $d$ is remaining constant at all, the term approaches 0. However, in opposite situation when $d$ increases large and growth rate of $N$ is not as fast, the $\frac{d}{0.05*N}$ approaches $+\infty$. Assuming that a dataset cannot contain negative or 0 features or sample:

$$D: \{(d, N) \mid d \in (0, \infty), N \in (0, \infty)\}$$
$$R: \{f(d, N) \mid f(d, N) > 0\}$$

(1)

Range from equation 1 shows that value of $f(d, N)$ has no limit, meaning factor could boost initial metric such that it becomes greater than 1, therefore sigmoid function is implemented to compress all possible extreme outputs to range from 0 to 1.

$$f(d, N) = \frac{1}{1 + e^{-\left(\frac{d}{0.05*N} - 1\right)}} \qquad (2)$$

Subtracting 1 from the main term enables sigmoid function to be centered around 0.5. It is the balancing point that ensures the model is evaluated neutrally under optimal conditions when $\frac{d}{0.05*N}$ equals exactly 1. This adjustment is designed not to penalize the model in favorable conditions, but rather to apply a boost only when conditions are unfavorable, such as when the dataset size is small relative to high dimensionality. Hence, additional constraint was applied to the formula.

$$f(d, N) = \frac{1}{1 + e^{-\left(\frac{d}{0.05*N} - 1\right)}} - \frac{1}{1 + e^0} \qquad (3)$$

This makes sure that the model in optimal conditions gets 0 boost and better context goes beyond 0, but this is handled using max () function. Significantly, adverse conditions enhance the model's performance by maximum of 0.5, that leaves place for future variables to assist evaluating model's efficacy relative to other dataset characteristics. To summarize, the final form of feature dimensionality handling factor is:

$$f(d, N) = 1 + \max\left(0, \frac{1}{1 + e^{-\left(\frac{d}{0.05N} - 1\right)}} - \frac{1}{1 + e^0}\right) \qquad (4)$$

Equation 4 satisfies cases mentioned in Table 1 by boosting the model in overfitting risks and neither penalizing nor awarding model in other stable, optimal and good performance circumstances.

Class Imbalance

Class imbalance is a challenge that became more acknowledged with a growth of big data. Uneven distribution of classes within a dataset leads to several problems with machine learning prediction algorithms' performance. One of them is accuracy paradox, when there is a huge class imbalance and model simply predicts majority class all the time, hence achieving a high classification accuracy [17]. The reason for this lies in accuracy calculation. It computes the ratio of all correct predictions to total number of predictions, including forecasting both minority and majority classes [18]. As a result, normalized metric should handle scenarios when dataset is imbalanced by penalizing accuracy value. As described in features dimensionality section, metric is designed to boost model's performance if it was trained in unfavorable conditions. However, it will be penalized for high class imbalance even though it can be counted as adverse circumstances. The explanation for this is that all other unfavorable conditions decrease accuracy, so a normalized metric increases it. However, an unequal number of samples in different classes usually leads to an unrealistic growth in accuracy, so the metric will reduce accuracy in this case. Handling this problem is crucial due to the common presence of non-balanced datasets in the real world. For example, according to Satyendra Singh Rawat [19], fraud detection, spam detection, and software defect prediction applications commonly encounter unbalanced datasets because of the rapid growth of big data, which makes it difficult to lower class disproportionality. Class imbalance ratio is included in another factor in the formula:

$$Class\ Imbalance\ Ratio(CI) = \frac{Number\ of\ Majority\ Class\ Samples}{Number\ of\ Minority\ Class\ Samples} \qquad (5)$$

Equation 5 depicts how many times there are more samples in one class than in another. For instance, a model that only predicts the majority class could reach more than 75% accuracy in a dataset containing 75% majority class and 25% minority class without learning any significant patterns in a dataset. The class imbalance ratio (CI), which is the ratio of majority to minority class samples, can be used to penalize the accuracy metric in order to account for this. The majority class is three times as frequent as the minority class, for example, if CI = 3. The observed accuracy is normalized by dividing by $1 + \log(CI)$, which lowers the inflation brought on by forecasting the majority class. In this case, $\log(CI)$ serves as "information bias" that represents the strength of the imbalance. Information theory frequently uses the logarithmic function to scale values based on their relevance and CI does not show the probability of choosing one of the classes itself, but how many times the model is more likely to choose one class instead of another, in reference to the concepts of Shannon's entropy and information gain [20]. Adding 1 to the denominator prevents it from becoming zero in cases of perfect class balance. When classes are evenly distributed (CI = 1), $\log(CI)$ equals zero, which would otherwise make the denominator zero. This adjustment ensures the formula remains defined, even when there is no class imbalance. Another advantage of using $1 + \log(CI)$ in the denominator is a flexible adjustment that remains reasonable across a wide range of class imbalance ratios, from mild to extreme. Even in extremely unbalanced situations, the correction progressively reduces accuracy as CI rises without causing significant reductions. For instance, when there is a significant imbalance, such as 1000:1 (where CI = 1000), the calculation produces $1 + log(1000) = 4$. This preserves relevant accuracy values even under extreme class distributions by resulting in a slight adjustment rather than a severe penalty. As a result, the $1 + \log(CI)$ strategy provides a gradual, controlled decrease in accuracy, maintaining realism and interpretability across a variety of imbalances without excessively punishing in high CI scenarios. To sum up, the adjustment function that scales based on class imbalance is:

$$h(CI) = 1 + log(CI) \tag{6}$$

Where:

Class Imbalance Ratio (CI) = $\dfrac{Number\ of\ Majority\ Class\ Samples}{Number\ of\ Minority\ Class\ Samples}$

Moreover, classification machine learning models can encounter multiclass datasets. Binary classification cases identify imbalance degree by simply calculating the ratio of majority class to minority one, therefore regular class imbalance ratio (CI) is not suitable for multiclass cases. Research performed by Ravid Shwartz-Ziv [21] proposes that averaging ratios between each class and majority class is the effective way to capture the dataset's overall imbalance. Instead, average class imbalance ratio (ACIR) is introduced. When a dataset contains only two classes, the majority class's ratio to the minority one is commonly considered; however, ACIR takes into account each class's correlation to the majority class. After that, all the probabilities are added up and their average value is calculated. The final output is restricted to a range of 0 to 1, where 0 represents a very spread and unbalanced dataset and 1 represents a completely balanced dataset. It makes the average class imbalance ratio suitable to be included in the formula.

$$ACIR = \frac{1}{C} \sum_{i=1}^{C} \frac{Size\ of\ class\ i}{Size\ of\ majority\ class} \tag{7}$$

Considering the logic of normalized metric in case of binary classification problems, low ACIR value, meaning extreme imbalance across classes has to penalize the model's performance. Meanwhile, a high ACIR value must not either penalize or boost the metric's value. Therefore, the concluding format of average class imbalance ratio is:

$$h(ACIR) = 1 + log\left(\frac{1}{ACIR}\right) \tag{8}$$

As mentioned earlier, this metric will be tested on several models, each suitable for a specific type of task. Support vector machine (SVM), K-means, and logistic regression models will demonstrate classification, clustering, and regression, respectively. The metrics considered before are suitable for both types of classification, both binary and multi-class. Logistic regression will also predict the probability of any samples belonging to one of the two classes, so the class imbalance ratio (CI) is suitable for it. Besides, the imbalance problem exists in clustering machine learning models as well. Clustering algorithms frequently produce uneven distribution as a result of issues like overlapping data, high-dimensional spaces, and sensitivity to initial parameters. Clusters of varying sizes may result from these circumstances, with some samples remaining underrepresented and others having a majority of the samples. The algorithm's attempt to reduce within-cluster variance, particularly in data without distinct separation or natural grouping structures, causes this imbalance. According to research carried out by Yudong He [22], because there is an absence of labeled data in unsupervised learning tasks, cluster sizes can fluctuate greatly and unpredictably. Class imbalances remain even though there are no designated classes since the majority of samples might have identical characteristics while others vary widely. Cluster imbalance may arise as a result, with larger clusters forming around recurring patterns and smaller clusters forming around unique cases. Since there is no initial label-based structure to guide modifications, class imbalance in clustering only becomes noticeable after the model has completed assigning clusters, in contrast to supervised learning where it may be controlled using resampling or boosting. Therefore, the normalized metric will penalize model if clusters differ significantly by modifying average class imbalance ratio (ACIR $_{class}$) to average cluster imbalance ratio (ACIR $_{cluster}$). Hence, cluster imbalance factor for clustering algorithms is:

$$h(ACIR_{cluster}) = 1 + log(ACIR_{cluster}) \tag{9}$$

where:

$$ACIR_{cluster} = \frac{1}{C}\sum_{i=1}^{C} \frac{Size\ of\ cluster\ i}{Size\ of\ majority\ cluster}$$

Signal-to-Noise ratio

Signal-to-Noise ratio is another concept commonly used to compare the degree of useful signal to background noise [23]. This measurement is exploited in communication systems, audio systems, and radar systems; however, the logic that lies behind this concept can be referred to as the machine learning domain as well. In this scenario, signal is associated with the relevant and meaningful information, while noise is the irrelevant and random data, such as outliers [24]. As a result, the Signal-to-Noise ratio (SNR) represents the quality and reliability of information within a dataset, measured in decibels. It fits into the concept of the normalized metric because the model fed on the small dataset probably results in a vast amount of noise and low accuracy (signal). With the increase in dataset size, noise level decreases and the model is expected to learn more meaningful patterns from it. Recently, research carried out by Tongtong Yuan [25] has found the positive effect of using SNR in clustering-related problems. It replaced the traditional Euclidean distance used to determine how close data vectors are to cluster centroid with the Deep SNR-based Metric Learning (DSML) metric derived from basic concepts of signal-to-noise ratio (SNR) from the engineering field. One way to use the available dataset characteristics and convert them to SNR is shown using the following formula [26]:

$$SNR = 10 * \log_{10} \frac{\sum y_{test}^2}{\sum (y_{pred} - y_{test})^2} \quad (10)$$

where:

$y_{test}$: Represents the true or actual output values. These are the correct values used as a benchmark to measure performance.

$y_{pred}$: Represents the predicted output values from the model. These are the values produced when attempting to replicate or denoise the input signal.

The logarithmic function is used to compress the output as a final result might vary significantly. Still, even after implementing a logarithm result number might vary greatly, thereby the standardized measurements are used to scale the SNR value from 0 to 0.5, as the feature dimensionality factor from the section above is located on such scale as well [27].

| SNR Range | Signal (Data) Quality |
|---|---|
| More than 40 dB | Excellent signal |
| 25 dB to 40 dB | Very good signal |
| 15 dB to 25 dB | Low signal |
| 10 dB to 15 dB | Very low signal |
| Less than 10 dB | No signal |

Table 4 - SNR Signal (Data) Quality Classification

The described data quality classification, derived from useful and non-relevant information from the model's results, is used to build a piecewise function that compresses the wide range of SNR values to a fixed 0 to 0.5 scale.

$$Normalized\ SNR(x) = \begin{cases} 0.125 + 0.125 * \frac{x - 0}{10}, & for\ 0 \leq x < 10 \\ 0.25 + 0.125 * \frac{x - 10}{5}, & for\ 10 \leq x < 15 \\ 0.375 + 0.125 * \frac{x - 15}{10}, & for\ 15 \leq x < 25 \\ 0.5 + 0.125 * \frac{x - 25}{15}, & for\ 25 \leq x < 40 \\ 0.5, & for\ x \geq 40 \end{cases} \quad (11)$$

Considering the possibility of normalized SNR being 0, a value of 1 is added to the final version of signal to noise ratio factor:

$$g(SNR) = 1 + SNR_{normalized} \tag{12}$$

Given the context of the research, a normalized metric should be effective for different types of tasks, such as classification, regression, and clustering. Therefore, equation 12 is not suitable for all cases and must be modified separately for each situation. The equation above is suitable for regression problems, but the context of classification and clustering is very different, and the SNR can still be calculated. For binary classification problems, signal-to-noise ratio equation is:

$$SNR_{binary} = 10 * \log_{10} \frac{\Sigma(y_{test} == y_{pred})}{\Sigma(1 - y_{prob})^2} \tag{13}$$

where:

$y_{test}$ is the true label of each sample, representing the actual class

$y_{pred}$ is the label predicted by a model for each sample, representing the class that the model predicts

$y_{prob}$ is the predicted probability for each sample, representing the probability that the sample belongs to predicted class

Here, the signal in this estimation of the Signal-to-Noise Ratio for the binary classification problem is the count of correct predictions, measured as $y_{test} == y_{pred}$. Since correct predictions directly reflect appropriate capturing of the pattern by the model in the data and, therefore, are a measure of the "strength" or "clarity" of its output. The higher the accuracy, as given by more correct predictions, the more signal there is. $(1 - y_{prob})^2$ represents the probability of the predicted class. High-confidence predictions have probabilities closer to 1 and can be regarded as less noisy due to the strong certainty of the model in its prediction. Low-confidence predictions are closer to 0 and hence noisier, in that it carries less reliability in the model output. Squaring $(1 - y_{prob})$ emphasizes the impact of these lower-confidence predictions, penalizing them more heavily as noise. Putting it altogether, this agrees rather well with notions of SNR from information theory: signal refers to desired clarity-correct predictions-and noise refers to undesirable variability, or uncertain predictions. In other words, this method basically quantifies the clarity of a model's performance on a classification task by comparing the strength of the correct classifications against the variability introduced by the uncertain predictions.

Considering the problem involving multiclass classification task, equation 13 derived for binary classification approach is also not suitable. As a result, the signal and noise in this instance will be different. The Signal in our SNR formulation is derived from the confusion matrix, which summarizes the counts of correctly and incorrectly classified samples across all classes. For multiclass classification with $C$ classes, the confusion matrix $M$ is a $C \times C$ matrix where each entry $M_{ij}$ represent the true positives for each class. To compute the signal, the squared values of these true positive counts are summed:

$$Signal = \sum_{i=1}^{C} M_{ij}^2 \quad (14)$$

By squaring each count, all classes in the total measure are still taken into account but classes with high correct counts and good predictive performance are highlighted. This method emphasizes instances when the model performs very well for particular classes and captures the model's capacity to predict each class consistently and accurately. The degree of uncertainty or mismatch between the actual class labels and the model's presented probability is reflected in the Noise term. The model generates a probability distribution across all classes for every sample. The "ideal" probability distribution for a sample $k$ with true class label $i$ is a vector where the probability for the true class $i$ is 1 and 0 for all other classes. Let $p_{k,j}$ represent the model's predicted probability for class $j$ on sample $k$. The noise for a single sample $k$ is calculated as the squared difference between the predicted probabilities and this ideal distribution:

$$Noise_k = \sum_{j=1}^{C} (p_{k,j} - \delta_{ij})^2 \quad (15)$$

where $\delta_{ij}$ is the Kronecker delta, equal to 1 if $j = i$ (the true class) and 0 otherwise [28]. Summing this across all samples in the test set gives the total noise:

$$Noise = \sum_{k=1}^{N} Noise_k \quad (16)$$

Using the computed signal and noise, the total SNR is calculated the same way as before:

$$SNR = 10 * \log_{10}\left(\frac{Signal}{Noise}\right) \quad (17)$$

Summary

The final general formula after accounting for all the prior factors that have an impact on the model's performance and can be used to assess the dataset's quality as well as the machine learning model's potential future performance on this particular type of dataset:

$$Normalized\ Metric = min\left(1, \frac{Performance\ Metric \cdot f(d,N) \cdot g(SNR)}{h(CI)}\right) \quad (18)$$

where:

- $f(d,N)$ represents the dimensionality adjustment based on dataset size $N$ and feature dimensionality count $d$,
- $g(SNR)$ is the Signal-to-Noise Ratio adjustment,
- $h(CI)$ is the class imbalance adjustment (or alternative $h(ACIR)$ imbalance measure for clustering and regression tasks).
- *Performance Metric can vary in relation to task type.*

Some factors are interpreted for each problem type specifically as mentioned above. For example, for binary classification, each component is further defined as:

$$\text{Normalized Metric} = \min\left(1, \frac{\text{Accuracy} \cdot \left(1 + \max\left(0, \frac{1}{1+e^{-\left(\frac{d}{0.05 \cdot N} - 1\right)}} - \frac{1}{1+e^0}\right)\right) \cdot \left(1 + SNR_{normalized}\left(10 \cdot \log_{10} \frac{\Sigma(y_{test} == y_{pred})}{\Sigma(1 - y_{prob})^2}\right)\right)}{1 + \log \frac{N_{majority}}{N_{minority}}}\right) \quad (19)$$

Where:

$$SNR_{normalized}(x) = \begin{cases} 0.125 + 0.125 * \frac{x - 0}{10}, & \text{for } 0 \leq x < 10 \\ 0.25 + 0.125 * \frac{x - 10}{5}, & \text{for } 10 \leq x < 15 \\ 0.375 + 0.125 * \frac{x - 15}{10}, & \text{for } 15 \leq x < 25 \\ 0.5 + 0.125 * \frac{x - 25}{15}, & \text{for } 25 \leq x < 40 \\ 0.5, & \text{for } x \geq 40 \end{cases}$$

Regression, clustering, binary classification, and multiclass classification are the four categories of machine learning tasks for which evaluations will be carried out independently in order to assess the efficacy of the proposed metric. Support Vector Machines (SVM) will be applied to classification tasks, K-means clustering to clustering activities, and Linear Regression to regression tasks. Datasets from the UCI official repository will be used to feed these models. Python will be used for both implementation and testing. The key objective is to show that, despite variations in dataset size, the adjusted measure continuously stays near a number that represents the model's actual potential. The Mean Absolute Deviation (MAD) will be computed in order to measure this stability.

**Results**

Binary Classification

A dataset containing information on loan approvals was selected to test the normalized metric for tasks related to binary classification [29]. There are 13 features in the dataset excluding one as a target variable. Based on the conclusions concluded earlier, for such a dataset, the optimal point of balance between features and the size of the dataset will be 260 samples. The accuracy and normalized accuracy were calculated between 80 and 1000 samples.

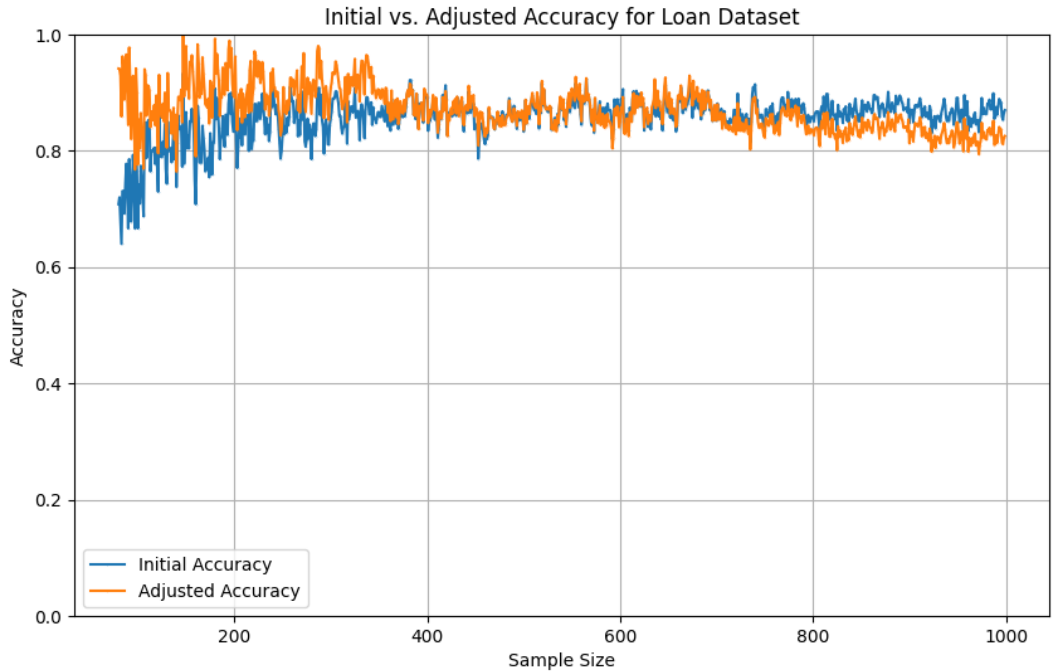

Figure 10 - Comparison of Initial and Adjusted Accuracy Across Varying Dataset Sizes

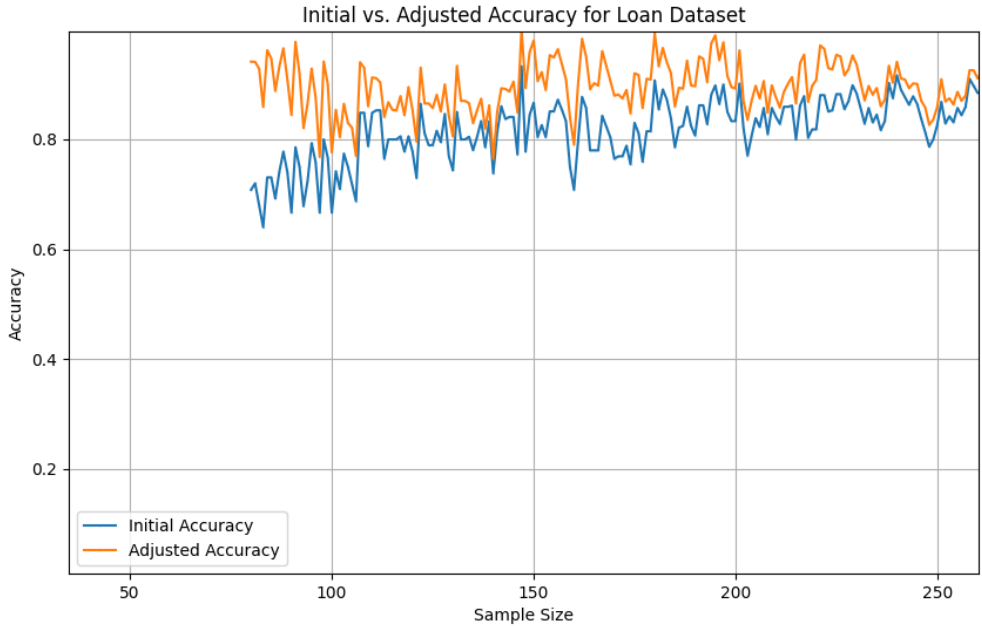

Figure 11 - Initial and Adjusted Accuracy vs. Training Dataset Size (Zoomed to 260 Samples)

The figures show how initial and adjusted accuracy on the loan dataset behaves as the sample size grows. When sample numbers are small, initial accuracy exhibits notable fluctuations, indicating how sensitive the model is to shifts in the data. Conversely, even with fewer samples, adjusted accuracy gets closer to its stable value, indicating the metric's capacity to forecast the model's possible performance immediately. Both metrics stabilize as sample sizes increase, although adjusted accuracy shows more constancy across sample sizes. As more data is supplied, the performance curve becomes smoother because of the updated metric's ability to effectively account for variables including sample size, signal-to-noise ratio (SNR), and class imbalance.

| Metric | Overall Average | Before 260 samples | After 260 samples | MAD from distance |
|---|---|---|---|---|
| Initial Accuracy | 0.857 | 0.816 | 0.867 | 0.0241 |
| Adjusted Accuracy | 0.871 | 0.895 | 0.864 | 0.0231 |

Table 5 - Comparison of Average Initial and Adjusted Accuracy Before and After 260 Samples

The initial accuracy and adjusted accuracy metrics are compared in the table before and after a threshold of 260 samples. The adjusted accuracy, which averages 0.895 for smaller sample sizes (before 260 samples), is significantly closer to its steady performance under ideal circumstances than the initial accuracy, which averages 0.816. Initial accuracy averages 0.867, while normalized accuracy stays steady at 0.864 when sample sizes exceed 260. This suggests that the adjusted metric has already predicted the model's possible performance. More data on stability can be discovered in the Mean Absolute Deviation (MAD) from the target; adjusted accuracy has a lower MAD (0.0231) than Initial Accuracy (0.0241). Greater consistency is suggested by the normalized metric's smaller MAD, which shows that it not only accurately predicts the model's eventual performance level but also does so with fewer fluctuations, making it a more trustworthy metric under different dataset conditions.

Multiclass Classification

Multiclass classification task differs from the binary one, described above. The machine learning model has to predict more classes, meaning finding more patterns in the available data. Moreover, the class imbalance factor affects more as the number of classes increases. As a result, measuring efficacy of the normalized metric at evaluating model's potential and performance within the multiclass domain is crucial. A dataset, consisting of 24 features and 3000 data entries, about comprehensive collection of health, lifestyle, and demographic information [30]. However, only half of the samples has been utilized as such portion can be count as "big" dataset and justified by this number being far away from the optimal point of 480 samples.

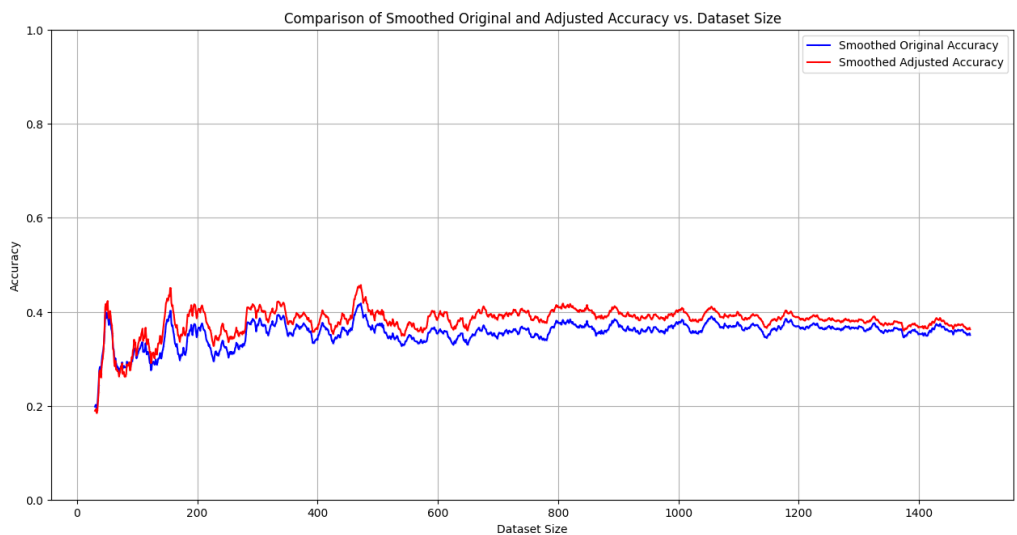

Figure 12 - Comparison of Smoothed Original and Adjusted Accuracy for Multiclass Classification vs. Dataset Size

Smoothed original and adjusted accuracy for multiclass classification across varying dataset sizes is compared in the first figure across a wider range of dataset sizes, including 1,500 samples. The adjusted (normalized) accuracy consistently tracks slightly above the original accuracy, indicating that the normalized metric effectively enhances performance predictions by factoring in dataset size, feature complexity, and other adjustments. This adjustment provides a more consistent, reliable depiction of the model's performance across a range of dataset sizes by reducing volatility. The normalized metric offers a trustworthy long-term performance estimate as additional data becomes available, as evidenced by the steady stabilization of both metrics as dataset size grows.

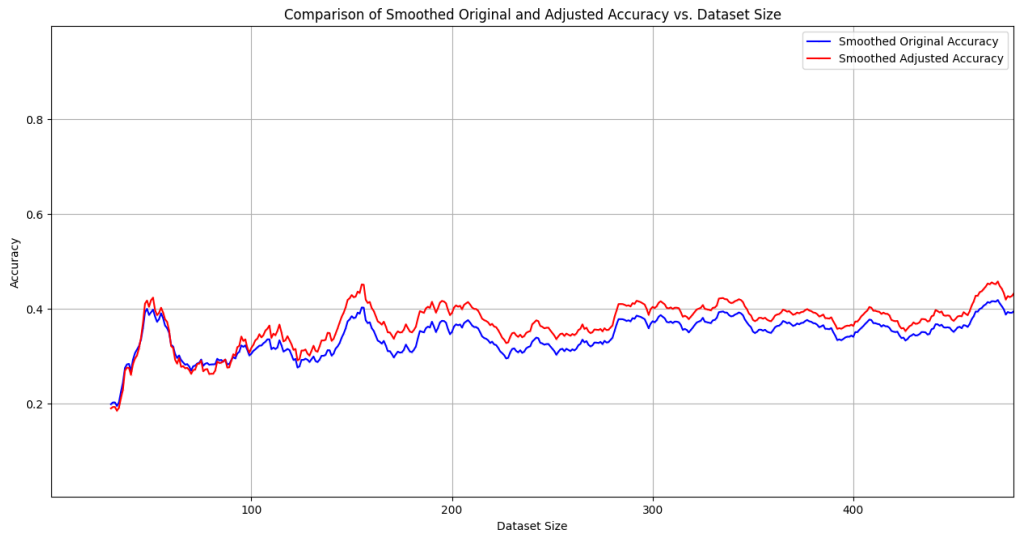

Figure 13 - Comparison of Smoothed Original and Adjusted Accuracy for Multiclass Classification vs. Dataset Size (Zoomed to 480 samples)

The behavior of the normalized metric at smaller dataset sizes is examined in greater detail in the second figure, which concentrates on the first range of dataset sizes (up to 480 samples). Even with a tiny dataset, the normalized (adjusted) accuracy in this zoomed view predicts higher values right away. This early increase in accuracy, which may be the result of the normalized metric's corrections for dimensionality and imbalance, shows how well it captures possible model performance early on. The normalized metric delivers a more optimistic but controlled measure of the model's potential by giving a greater accuracy estimate immediately. As more data becomes available, the metric stabilizes.

| Metric | Overall Average | Before 480 samples | After 480 samples | MAD from target |
| --- | --- | --- | --- | --- |
| Initial Accuracy | 0.355 | 0.339 | 0.362 | 0.023 |
| Adjusted Accuracy | 0.379 | 0.364 | 0.387 | 0.023 |

Table 6 - Comparison of Mean Initial and Adjusted Accuracy Before and After 480 Samples

Regression

A linear regression model was trained on the loan dataset from the UCI repository. Fluctuations between the original metric and the adjusted metric were observed by incrementally increasing the dataset size from 30 to 1500 samples. The dataset included 21 features, excluding the target column [31]. Based on prior calculations for the optimal sample size per feature count, 420 samples is considered the ideal threshold. Dataset sizes below this threshold present challenging conditions for the model, while sizes beyond 420 samples create more favorable conditions.

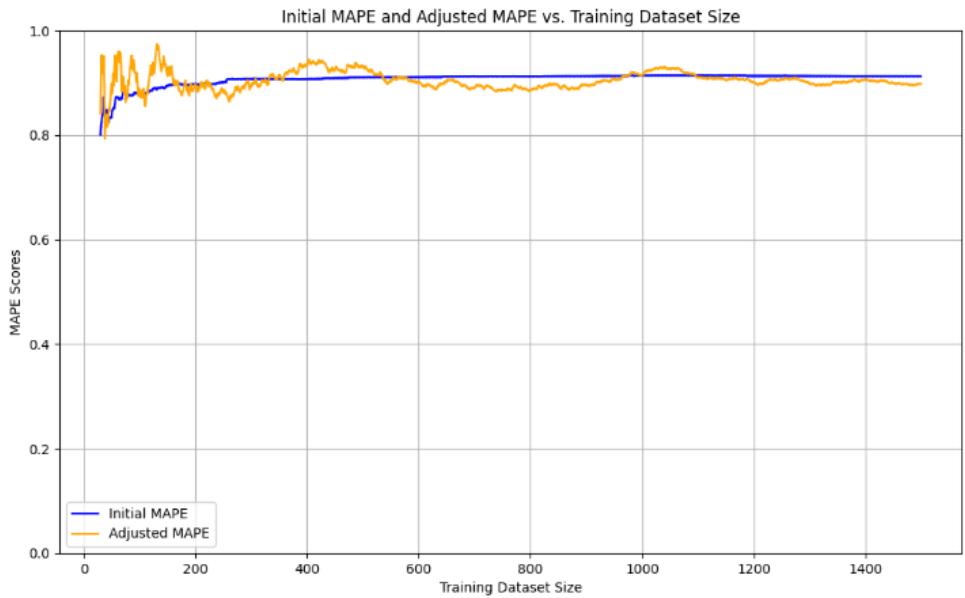

Figure 14 - Comparison of Initial and Adjusted MAPE Across Varying Dataset Sizes

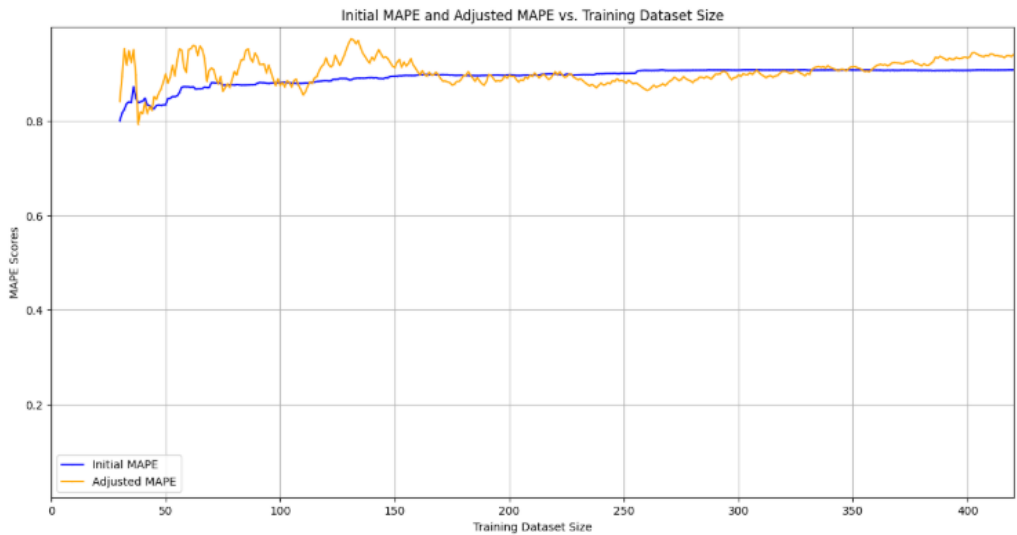

Figure 15 - Initial and Adjusted MAPE vs. Training Dataset Size (Zoomed to 420 Samples)

The comparison of Initial and Adjusted MAPE for different dataset sizes is shown in this figure. Like accuracy in classification, Mean Absolute Percentage Error (MAPE) is an accuracy indicator used for regression tasks. It calculates the average % error between the actual and predicted values; more accuracy is shown by lower numbers. To make the results more convenient to interpret, 1−MAPE values are displayed instead, so higher values now indicate better accuracy. As expected, the Initial MAPE (blue line) in the plot steadily improves as the dataset size grows. This pattern implies that the model learns better and generates more accurate predictions with more data, eventually stabilizing at about 420 samples. Even with small datasets, the Adjusted MAPE (orange line) starts approaching the values it presents with greater dataset sizes. This suggests that the model's potential performance is reflected in the adjusted metric early on, with consistent values that correspond to the model's final performance. Initially, there are minor fluctuations, but as the dataset size increases, these decrease and both metrics converge. This comparison shows that even in difficult situations with lesser datasets, the adjusted metric shows the model's potential immediately.

| Metric | Overall Average | Before 420 Samples | After 420 samples | MAD from Target |
|---|---|---|---|---|
| Initial MAPE | 0.908 | 0.894 | 0.913 | 0.019 |
| Adjusted MAPE | 0.905 | 0.904 | 0.905 | 0.001 |

Table 7 - Comparison of Average Initial and Adjusted MAPE Before and After 420 Samples

According to the data, the Adjusted MAPE is closer to the Initial MAPE after 420 samples than it is to the Initial MAPE before 420 samples. This implies that while the initial metric needs a larger dataset to achieve comparable levels of accuracy, the normalized metric more precisely represents the model's potential performance even with a smaller dataset. The Adjusted MAPE is substantially more consistent with the "future" target values (values after 420 samples) than the Initial MAPE, according to the Mean Absolute Deviation (MAD) results. In particular, the Adjusted MAPE's MAD is 0.0013, while the Initial MAPE's is 0.0187. A greater degree of stability and predictive reliability for the normalized metric over the original one is indicated by the lower MAD value for Adjusted MAPE, which shows that it stays considerably closer to the stable, long-term performance values even at reduced dataset sizes.

Clustering

The official "Wine" dataset from the UCI repository was exploited for the clustering task. The dataset contains 13 features and is perfectly suitable for the clustering task; however, it only contains 178 instances [32]. Therefore, it was expanded synthetically using the nearest neighbor interpolation approach, where synthetic samples are created by interpolating between existing samples.

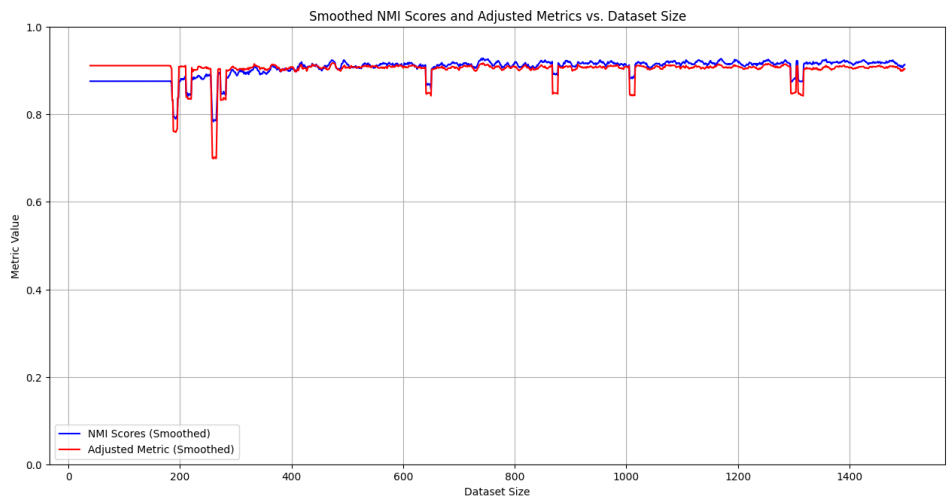

Figure 16 - Comparison of Initial and Adjusted NMI Across Varying Dataset Sizes

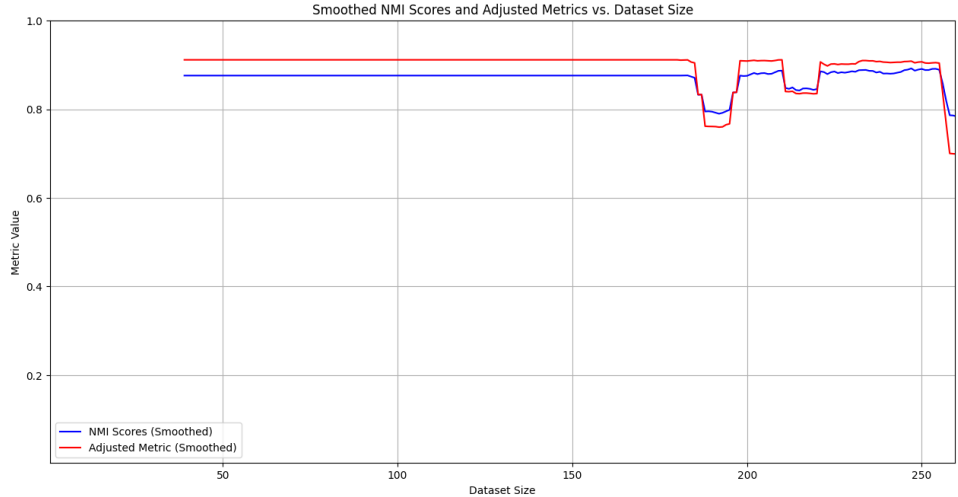

Figure 17 - Initial and Adjusted NMI vs. Training Dataset Size (Zoomed to 260 Samples)

As the dataset size grows, the chart compares the adjusted metric with smoothed normalized mutual information (NMI) Scores. The dependence between predicted clusters and true class labels is measured by the NMI, a metric used to evaluate the quality of clustering. Better clustering performance is indicated by larger NMI values, which range from 0 to 1. A value of 1 denotes perfect alignment between clusters and true labels. Because it offers a normalized metric that takes into consideration the unpredictability of clustering tasks, NMI is frequently used. Both metrics initially display variability in the graph, with the normalized metric in particular showing noticeable declines with smaller dataset sizes. The adjusted metric's sensitivity to variables like noise and class imbalance, which can have a big impact on clustering results in smaller datasets, is probably the cause of this variability. Both metrics show steady clustering performance as dataset size grows, stabilizing and converging. As dataset size increases, the Adjusted Metric nearly matches NMI scores and approaches stability faster than NMI. This suggests that the model's performance potential, even with smaller datasets, is reflected in the Adjusted Metric's ability to capture the underlying clustering quality and achieve its stable value early. The normalized metric's slight deviations at bigger sizes show that it is still sensitive to variations in the distribution of data, highlighting how responsive it is to even the smallest changes that could affect clustering results.

| Metric | Overall Average | Before 260 samples | After 260 samples | MAD from target |
|---|---|---|---|---|
| Initial NMI | 0.906 | 0.870 | 0.913 | 0.0184 |
| Adjusted NMI | 0.904 | 0.895 | 0.906 | 0.0126 |

Table 8 - Comparison of Average Initial and Adjusted NMI Before and After 420 Samples

The adjusted NMI averages about 0.895 before reaching 260 samples, which represents less favorable dataset conditions. Under the same circumstances, the initial NMI averages approximately 0.870. The initial NMI increases to 0.913 as the dataset size improves (after 260 samples). The modified NMI before 260 samples is notably 0.025 closer to this number than the initial NMI, indicating that the modified metric approaches the desirable performance level 2.38 times more closely. This shows that even with poor dataset conditions, the formula tends to approximately represent the model's future performance.

**Discussion**

The derived normalized metric was tested on various models selected for different types of tasks. More precisely, the metric effect has been measured for binary and multiclass classification, regression, and clustering. The metric showed the expected results from it, thereby confirming the null hypothesis. Indeed, factors such as class imbalance, feature dimensionality, and the quality of the model's response to interfering noise (signal-to-noise ratio) affect how the machine learning model is trained. In all cases, different metrics were used that were suitable for various types of tasks, but the normalized metrics managed to adapt to each of them effectively. The metric worked best with a linear regression model. In this case, the metric was the most stable, maintaining an almost constant value before and after the optimal point, determined by the number of features. The value remained around 0.904 and 0.905, while showing the lowest MOD equal to 0.001, meaning that the metric almost does not deviate from the average performance of the model. The metric performed the worst with data classification when it contained several classes, that is, not binary. Since the metric is expected to show in advance the maximum result that the machine learning model will show in the future, it was assumed that there would be a stable value at the beginning, which the initial accuracy would reach after. However, throughout the entire interval, the adjusted metric copied the behavior of the original metric. This can be seen in detail in Figure 13, where both lines characterizing the normalized and the usual metric are not just similar but identical in upward and downward trends. That is, the derived formula changes the value exactly a little higher. It is worth noting that the values calculated after prove that even in this case the metric reached the future value much earlier than the original metric, however, extreme and visually striking differences are not noticeable, as in comparison with other models and graphs. The normalized metric's high reliance on the task's particular context is one of its limitations. Because imbalance and signal-to-noise ratio can be interpreted differently depending on the task, the formula has to be modified accordingly. For instance, the definitions of signal and noise in clustering and regression tasks differ. Because of this, even if a standard version of the metric has been developed, it is not universal for all kinds of tasks. Another drawback is the occurrence of extreme cases where the signal is significantly high, coupled with a high feature dimensionality and a limited sample size. Theoretically, the corrected accuracy in these situations may be greater than 1, which is unacceptable. This was fixed by adding a "min()" function to cap the measure at 1. Even though this scenario hasn't happened in practice, the theoretical possibility remains unmitigated in the calculations. Furthermore, datasets containing up to 1,500 samples were used to evaluate the models. Although this range was theoretically adequate to illustrate the normalized metric's behavior, further research might involve extending it to bigger dataset sizes to track any additional metric changes. However, at these bigger sizes, it is anticipated that the metric will be similar to traditional evaluation metrics like original accuracy, MAPE, or NMI. Only traditional machine learning methods, such as support vector machines, k-means, and linear regression, were used in the study. Testing the normalized metric on more advanced algorithms, such as neural networks, and applying it to evaluation metrics commonly used in those contexts could provide valuable insights. It would be beneficial to look into how the normalized metric functions in these scenarios because complex models might be impacted by other dataset properties. Even with small or fluctuating data sets, the normalized metric offers encouraging prospects for real-world applications in various fields where reliable performance evaluation is essential. Because of privacy and data-sharing limitations, machine learning models in healthcare diagnostics frequently work with small or spread datasets. Particularly in the early phases of deployment, models that forecast patient diagnoses, treatment outcomes, or risk factors require trustworthy accuracy metrics to guarantee that the outputs are legitimate. By accounting for the small data sizes and high feature dimensionality common in healthcare data, such as test results, genetic information, or patient history, the normalized metric may offer a more consistent performance evaluation. When data is limited or steadily growing, the metric may help with clinical decision-making and improve model reliability by providing more consistent performance indicators. Risk models in the financial industry usually use sparse datasets about small businesses, emerging markets, or new financial products. By offering a trustworthy performance evaluation that balances high feature complexity and takes sample size fluctuation into account, the normalized metric may be useful in this field. The adjusted metric provides a useful method for comprehending model behavior early on and forecasting long-term reliability, for example, in loan risk assessments or fraud detection, where model predictions have a direct influence on financial decisions. This could help financial institutions manage risk more effectively by empowering them to make well-informed decisions even in the face of inadequate data. Machine learning models are being utilized more and more in e-learning and education applications to forecast learning outcomes, customize learning materials, and evaluate student performance. Data accessibility, however, can differ greatly among schools or groups of students. By evaluating model performance consistently across these different data sizes, the normalized metric can offer a more reliable accuracy indicator for performance predictions or personalized suggestions. For instance, in adaptive learning systems, even with small datasets, the metric's capacity to provide consistent performance estimates can be crucial in developing early-stage models that customize content according to a student's performance profile. Since datasets are frequently few in the early phases of model building, quick assessment is crucial to guiding future work. The normalized metric is a useful tool for preliminary model evaluations since it may accurately forecast model performance even with sparse data. The metric aids in prioritizing models that are expected to perform well with additional data by offering insights into a model's potential stability and efficacy early on, maximizing resources during the model-building process. This method facilitates more effective decision-making in model selection and improvement, which is advantageous across industries where rapid evaluation and prototyping are essential.

**Conclusion**

The research presented a normalized, dataset-adaptive metric for assessing the performance of machine learning models with different dataset complications. The measure provides a more comprehensive and scalable alternative over traditional evaluation metrics by effectively including important dataset aspects like size, feature dimensionality, class imbalance, and noise. Extensive testing on regression, clustering, and binary and multiclass classification tasks showed that the normalized metric could effectively adjust to various settings, yielding solid and accurate estimates even in the face of unfavorable conditions. Practitioners in related fields were surveyed to get their opinions on the effectiveness and design of the normalized metric. The findings showed a high degree of agreement with the metric's stated objectives, and respondents acknowledged its capacity to offer reliable and useful performance insights. The survey also confirmed the metric's applicability in solving actual machine-learning problems. The metric's capacity to accurately forecast model potential from limited data, showing future performance trends that traditional metrics tend to overlook in the early stages, is one of the study's major accomplishments. This feature has major implications for real-world applications, especially in domains where data variability and scarcity are frequent problems, such as healthcare, financial risk modeling, education, and early-stage prototyping. The metric's usefulness in providing consistent performance evaluation is further demonstrated by the fact that it stabilizes earlier than traditional measures. Despite its advantages, the study identified certain drawbacks. The metric is highly dependent on the particular task setting, requiring changes for variations in imbalance and noise definitions between tasks such as regression and clustering. Furthermore, extreme situations with high signal-to-noise ratios and feature dimensionality with small sample sizes exposed theoretical issues that were resolved by capping the metric values. These problems highlight the necessity of additional improvement to increase the metric's robustness and universality. The impact of a normalized, property-based metric on assessing machine learning models under various dataset limitations was the research question of the study, and it was thoroughly answered. The results showed that the normalized metric predicts scalability as data availability rises and offers more consistent evaluations. Future research could investigate the metric's applicability to advanced machine learning models like neural networks and broaden the testing range to larger datasets. The usefulness and applicability of the normalized metric will be further confirmed by examining how it interacts with complex data aspects and task-specific difficulties. In the end, the normalized measure addresses the drawbacks of traditional metrics and offers the possibility of more adaptive and effective model development by providing a useful tool for assessing the model's effectiveness in real-world scenarios.